\documentclass[conference]{IEEEtran}
\IEEEoverridecommandlockouts
\usepackage[T1]{fontenc}
\usepackage[utf8]{inputenc}
\usepackage{amsmath,amssymb}
\usepackage{graphicx}
\usepackage{booktabs}
\usepackage{cite}
\usepackage{xcolor}
\usepackage[hidelinks]{hyperref}
\usepackage{url}
\usepackage{array}
\newcolumntype{P}[1]{>{\raggedright\arraybackslash}p{#1}}

\newcommand{\orcid}[1]{\href{https://orcid.org/#1}{ORCID: #1}}

\begin{document}

\title{Compact and Decodable Electrocardiogram Text for Large Language Models\thanks{Preprint. This paper is a conference-format version of a chapter submitted to the edited volume \emph{Advances in Artificial Intelligence and Data Science: Methods, Systems, and Applications} (Karab\"uk University Press).}}

\author{\IEEEauthorblockN{Elif Doruk\-ba\c{s}{\i}}
\IEEEauthorblockA{\textit{Department of Computer Engineering}\\
\textit{Faculty of Computer and Information Sciences}\\
\textit{Karab\"uk University}, Karab\"uk, T\"urkiye\\
elifkabullar@karabuk.edu.tr\\
\orcid{0000-0003-3206-5784}}
\and
\IEEEauthorblockN{\.Ilker T\"urker\textsuperscript{*}}
\IEEEauthorblockA{\textit{Department of Computer Engineering}\\
\textit{Faculty of Computer and Information Sciences}\\
\textit{Karab\"uk University}, Karab\"uk, T\"urkiye\\
iturker@karabuk.edu.tr\\
\orcid{0000-0001-7577-4658}}
}

\maketitle

\begingroup\renewcommand\thefootnote{}\footnotetext{\textsuperscript{*}Corresponding author: \.Ilker T\"urker (iturker@karabuk.edu.tr).}\endgroup

\begin{abstract}
When a twelve-lead electrocardiogram is written as text for a large language model, the number of tokens has to be balanced against the information lost through compression. This paper presents a deterministic encoding that writes eight independent leads as amplitude letters and exact time gaps and recovers the remaining four limb leads algebraically. The key points are selected with a cardiac-cycle-dependent tolerance, the amplitudes are parameterized by a 26-letter logarithmic codebook learned from a training set, and a 10-step tolerance schedule keeps all records within a fixed token budget. The encoding was applied to 21,388 records from the public PTB-XL database and evaluated using an exact decode-and-re-encode round trip, waveform reconstruction error, and a paired comparison of convolutional classifiers trained with 5 different seeds on signals from the original and decoded databases. All records were contained in 4,080 tokens (average: 2,866 tokens, 64.3\% fewer than a per-sample transcription), and all stored lead strings were exactly reproduced. The average correlation between the decoded and reference signals was 0.982. The macro-averaged area under the receiver operating characteristic curve was 0.9167 for the reference signals and 0.9157 for the decoded signals (paired difference $-$0.0010; 95\% patient-level bootstrap interval, $-$0.0028 to 0.0007), within the pre-specified non-inferiority margin of $-$0.010; this is not a confirmatory result, because the test fold had been used during encoder development. The diagnostic content of a compact text can therefore be preserved, while the language model's ability to read the text is yet another matter.
\end{abstract}


\section{Introduction}
The electrocardiogram (ECG) is among the most frequently recorded clinical signals, and its 12 standard leads capture the heart's electrical activity from complementary perspectives. Large language models (LLMs) are being explored as tools that could combine such recordings with clinical text, answer questions about them, or draft reports \cite{r1}. To do so, the waveform must first reach the model in a form it can process. In one route, a dedicated signal encoder produces embeddings that are aligned with the input space of a particular language model \cite{r2,r3}. In the other, the waveform is written as a sequence of discrete symbols that the model reads like text \cite{r4,r5,r6}.

Text can be inspected, stored, and passed to different models without a new alignment module, as ECG-aBcDe demonstrates by fine-tuning several language models on a single encoding while keeping their tokenizers unchanged \cite{r5}. Text also has a cost. A ten-second, twelve-lead recording sampled at 100 Hz contains 12,000 voltage values before lead labels, separators, and instructions are added; studies that prompt LLMs with physiological time series show why the numerical format of the input matters \cite{r7}; and the cost of standard self-attention grows quadratically with sequence length \cite{r8}. An encoding must therefore control the length produced by a real tokenizer while still covering the complete recording.

Evaluation is another challenge. The diagnostic score of an LLM trained on encoded ECGs combines two factors: the information that was not corrupted during encoding and the model's ability to effectively utilize the information it was fed. Recent research indicates that the representation of the input can cause the predictions of LLMs to shift \cite{r9} and that a small reconstruction error in the whole can mask the distortion of a short but clinically relevant deflection. What is missing is a protocol for quantifying the information content of an ECG representation in text, independent of any language model. ECG-26c fills that void by encoding each record deterministically as eight labeled lead lines, in which amplitude letters alternate with the exact time gaps between consecutive key points, under an explicitly specified token budget. It asks whether it is possible to record the entire signal without truncation given a fixed context length, how much waveform error is added by the encoding, and whether a classifier trained on the decoded signals will lose some diagnostic information relative to an identical classifier trained on the reference signals.

This paper makes the following contributions:

\begin{itemize}
\item An eight-lead text encoding that combines heartbeat-based key-point selection, a logarithmic amplitude codebook, exact time gaps, and a tolerance schedule that keeps every PTB-XL record within 4,080 tokens.
\item An evaluation protocol that separates codec consistency, waveform fidelity, and diagnostic information, using a paired 5-seed classifier design and patient-level bootstrap intervals.
\item An ablation over fixed compression levels that relates compression and reconstruction error to classifier performance and shows that reconstruction error is not a good measure of diagnostic loss.
\end{itemize}

Section~\ref{sec:related} provides the background and reviews related representations. Section~\ref{sec:method} describes the encoder and decoder, Section~\ref{sec:setup} the dataset and the evaluation protocol, and Section~\ref{sec:results} the results. Section~\ref{sec:conclusion} concludes the paper.

\section{Related Work and Background}\label{sec:related}
\subsection{Background}
\subsubsection{The ECG is a multichannel time series}
Leads I, II, and III and the augmented leads aVR, aVL, and aVF are obtained from three limb electrodes, whereas V1--V6 come from six chest electrodes. Because the six limb leads share the same electrodes, only two of them are independent; the other four follow from Einthoven's and Goldberger's relations. The deflections produced by each heartbeat are typical: P wave, atrial depolarization; QRS complex, ventricular depolarization; T wave, ventricular repolarization. The P and T waves are slower and smaller, and the QRS complex has the steepest changes in the recording and normally lasts less than 0.12 s. Timing is as important as amplitude; e.g., the PR interval has diagnostic value.

\subsubsection{Notation}
A recording is a matrix $X$ of 12 rows (leads) by 1,000 columns (10~s at 100~Hz), and $x_\ell[n]$ is the value of its $\ell$-th lead at sample $n$. An encoder $E$ maps $X$ to a string $s$, and a decoder $D$ maps it back to an approximation $\hat{X} = D(s)$. Three properties of the pair $(E, D)$ are explored. \emph{Consistency}: re-encoding the decoded key points gives back the original string $s$; \emph{fidelity}: $\hat{X}$ is close to $X$; \emph{sufficiency}: a model trained on $\hat{X}$ performs as well as one trained on $X$. Only consistency can be checked exactly; the others must be measured.

\subsubsection{Simplification and quantization}
Polyline simplification keeps only the samples needed to approximate a curve within a tolerance. The Ramer--Douglas--Peucker (RDP) algorithm does this recursively, keeping the point farthest from the current chord at each stage until all the deviations are less than the tolerance \cite{r10,r11}. Amplitude quantization then maps each retained value to one of a finite set of levels; logarithmic ($\mu$-law) companding, unlike uniform quantization, uses narrow bins near zero and wide bins at large magnitudes, which is well-suited to signals that cluster around a baseline. Both steps discard information, so a decodable representation is not lossless: decoding yields a single, deterministic approximation of the input.

\subsection{Related Studies}
\subsubsection{ECG--language models built on learned encoders}
A signal encoder can be linked to a language model via learned projections, as in \cite{r2,r3}, and the ECG representation can be pretrained from cardiological text using a language model, as in \cite{r12}; a recent survey puts these systems into context with the rest of the ECG literature \cite{r1}. They take advantage of abundant connections between waveforms and clinical terms. Still, the continuous embedding they pass to the language model cannot be read or decoded into a waveform, and the alignment must be repeated for every new model.

\subsubsection{Symbolic and learned vocabularies}
SAX maps segment averages to equiprobable symbols \cite{r13}, and LLM-ABBA approximates a series with adaptive linear pieces before symbolizing them for language models \cite{r14}. In ECGs, ECGBERT is trained using self-supervised learning to obtain representations from segments considered as words \cite{r15}; ECG-Byte quantizes the signal and learns byte-pair encoding (BPE) merges \cite{r16} on the resulting strings to get a waveform-specific tokenizer \cite{r4}; HeartLLM quantizes the lead-wise waveform and extends the vocabulary, and it pretrains the language model before instruction tuning \cite{r6}. Learned vocabularies are efficient to store; however, symbols in the vocabulary do not have an innate physical meaning, and the number of tokens depends on vocabularies that differ between studies.

\subsubsection{Fixed alphabets with physical meaning}
ECG-aBcDe selects breakpoints with $\ell$1 trend filtering \cite{r17}, writes the amplitude at each breakpoint as a lowercase letter and the elapsed time as an uppercase letter, and passes the text to unmodified language models \cite{r5}. Because each letter stands for a fixed bin, the text can be decoded into an approximate waveform. The published method, however, quantizes time gaps into equal-frequency bins, writes all twelve leads, and truncates inputs at 4,096 tokens, so the end of a long recording can be lost.

\subsubsection{Evaluating what a representation preserves}
The weighted diagnostic distortion measure was proposed because the percentage root-mean-square difference can remain small even when clinically important features are altered \cite{r18}. On PTB-XL, benchmark studies provide deep-learning baselines for the diagnostic superclasses \cite{r19}, and subsequent work reports that 100-Hz inputs perform as well as 500-Hz inputs \cite{r20}. These benchmarks evaluate classifiers, not representations.

\subsection{Research Gap}
The reviewed approaches are summarized in Table~\ref{tab:1}, which shows how ECG information is delivered to the language model, either via continuous embeddings, learned vocabularies, or fixed symbolic alphabets. The results of these studies do not show directly how much of the waveform information reaches the language model through these representations. The key to successful report generation or question answering is ensuring that the information in the input is correctly interpreted by the language model. The task of separating these two factors is still an important methodological problem.

ECG-aBcDe is relevant because each of its amplitude (lowercase) and duration (uppercase) letters stands for a fixed bin, so the ECG waveform can be reconstructed from the text. Its published evaluation focuses on question answering, and the authors also report timing distortions that arise during reconstruction, which they reduce by peak-alignment-based resampling \cite{r5}. The disadvantage of this, however, is that if time intervals are divided into equal-frequency bins, timing errors can add up, distorting or shifting the reconstructed waveform. One problem yet to be solved is how to shorten a full multilead recording while minimizing alterations to the sample locations of the retained key points. ECG-26c can overcome this problem by using a precise count of samples between key points. So it eliminates temporal quantization as a source of reconstruction errors, but still has lossy amplitude quantization and interpolation.

Another challenge is writing full recordings within a given token budget. There is no one-to-one mapping between sequence length (in characters or symbols) and the number of language-model tokens; if the sequence exceeds the model's context length, the rest is truncated. Length control should be applied during encoding and should use the tokenizer of the desired language model. The signal-text budget also has to be kept separate from the instructions and outputs during evaluation. When all recordings in a dataset fit within the signal-text budget, unseen recordings from the same source are likely to fit as well, but this is not guaranteed, and full downstream prompts still have to be counted separately.

Waveform similarity is also not enough to show that diagnostic information is preserved. Global reconstruction measures can mask short informative deflections \cite{r18}, and language model scores are sensitive to model capacity, training, and task formulation. A complementary evaluation is performed between classifiers trained with the filtered reference signal and the decoded signals, under matched data and training conditions. When assessing uncertainty in the paired performance difference, it is possible to evaluate the contribution of the encoding to a measurable loss in the diagnostic task in question.

ECG-26c combines all these requirements into a fixed-amplitude codebook with exact sample-gap encoding, plus a paired 5-seed classifier evaluation, and provides tokenizer-based length control. Before any language model is trained, its contribution is the joint evaluation of codec consistency, waveform fidelity, and diagnostic information. This evaluation is used to describe the amount of representation loss in the classifier family and the tested data set. It does not establish clinical equivalence or demonstrate that a language model can retrieve the retained information from the symbolic text.

\begin{table*}[t]
\caption{Comparison of related studies and the positioning of this paper.}
\label{tab:1}
\centering\footnotesize
\begin{tabular}{@{}P{3.0cm}P{1.6cm}P{4.1cm}P{3.6cm}P{3.6cm}@{}}
\toprule
Approach & Studies & Input to the language model & Meaning of symbols & Waveform recoverable from the input\\
\midrule
Encoder alignment & \cite{r2,r3,r12} & Continuous embeddings & No symbols & No \\
Generic symbolic time series & \cite{r13,r14} & Symbols for aggregates or linear pieces & Value bins or clustered pieces & Approximately \\
Learned ECG vocabularies & \cite{r4,r6,r15} & Learned tokens or codes & Learned from data & Depends on the codebook \\
Fixed ECG alphabet & \cite{r5} & Amplitude and time letters, 12 leads & Amplitude and duration bins & Approximately; time quantized \\
ECG-26c (this paper) & -- & Amplitude letters and exact gaps, 8 leads, $\le$ 4,080 tokens & $\mu$-law bins in mV; exact sample counts & Approximately; key-point times exact \\
\bottomrule
\end{tabular}
\end{table*}

\section{Method}\label{sec:method}
\subsection{Overview of the Proposed Approach}
\begin{figure*}[t]
\centering
\includegraphics[width=0.82\textwidth]{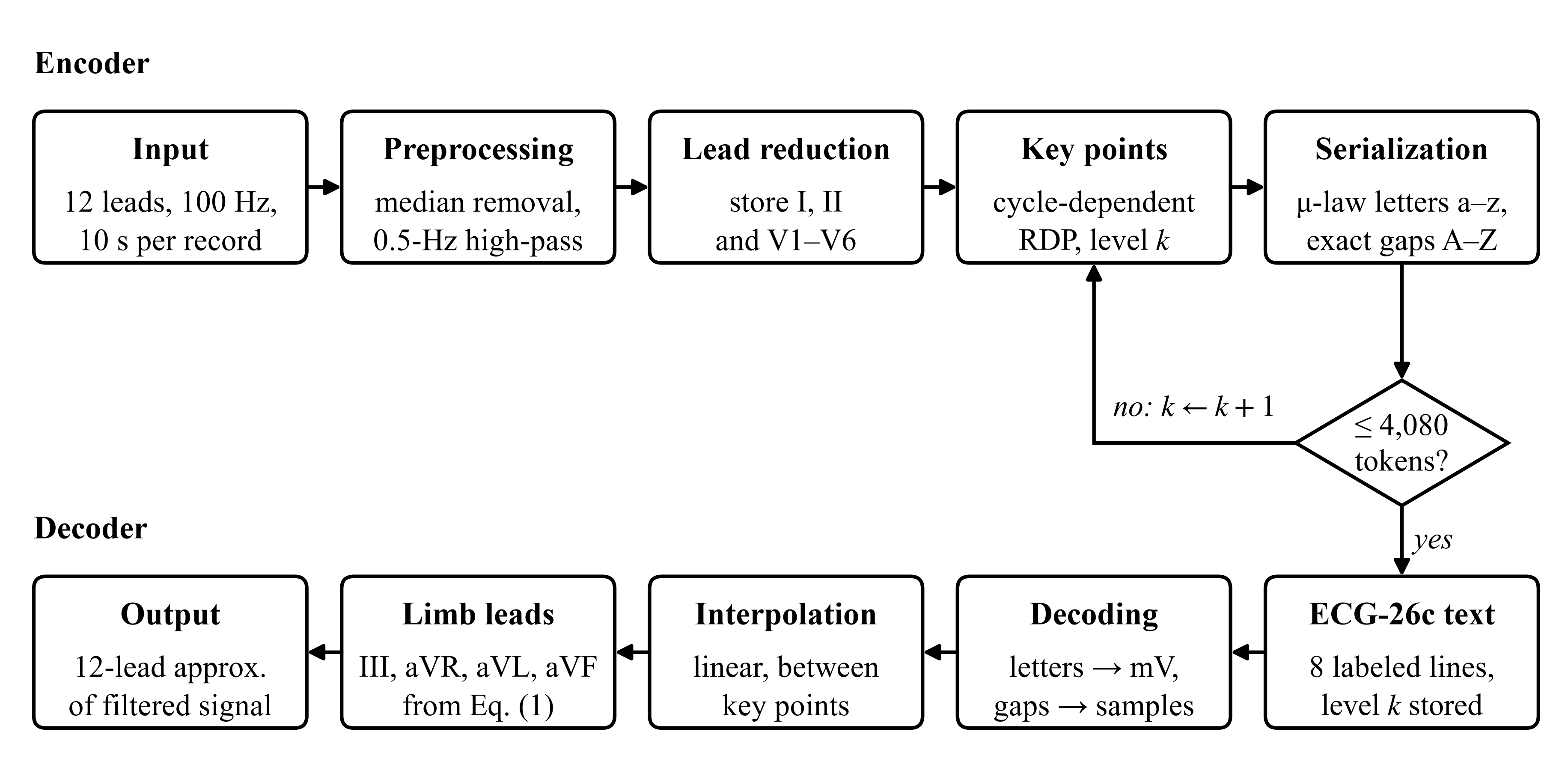}
\caption{Encoder (top) and decoder (bottom) of ECG-26c. The tolerance level k starts at 0 and increases only when the serialized text exceeds the token budget; decoding requires the frozen codebook and the format conventions but no learned model.}
\label{fig:1}
\end{figure*}

As Fig.~\ref{fig:1} shows, a record is first filtered and reduced to its eight independent leads. For each lead, key points are selected with tolerances that depend on their position within the cardiac cycle; their amplitudes are mapped to letters; and the gaps between them are written as exact sample counts. If the text exceeds the token budget, key points are selected again at a coarser tolerance level. The decoder reverses the serialization, interpolates between key points, and rebuilds the four omitted limb leads.

ECG-26c inherits from ECG-aBcDe the idea of alternating amplitude and time letters but changes most other design decisions (Table~\ref{tab:2}). The comparison concerns design, not performance, since the two methods were developed with different data and tasks.

\begin{table}[t]
\caption{Design comparison between ECG-aBcDe \cite{r5} and ECG-26c.}
\label{tab:2}
\centering\scriptsize
\begin{tabular}{@{}P{1.75cm}P{2.85cm}P{3.2cm}@{}}
\toprule
Design element & ECG-aBcDe & ECG-26c\\
\midrule
Sampling rate & Resampled to 250 Hz & 100 Hz (PTB-XL records100) \\
Leads written & All 12 & I, II and V1--V6; the other four rebuilt with Eq.~\eqref{eq:limb} \\
Key-point selection & $\ell$1 trend filtering & RDP with a cycle-dependent tolerance \\
Amplitude letters a--z & 24 equal-width bins between the 1st and 99th training percentiles, plus 2 open outer bins & $\mu$-law bins ($\mu$ = 20) between the 0.05th and 99.95th training percentiles; a--m $\le$ 0 mV $<$ n--z \\
Decoded amplitude & Upper bound of the bin & Mean of the training key points in the bin \\
Very large amplitudes & Absorbed by the outer letters & Lead written at half or quarter scale (x2, x4) \\
Time letters A--Z & 26 equal-frequency bins; one-sample gaps written & Exact gap in samples; one-sample gaps implicit \\
Length control & Truncation at 4,096 tokens & At most 4,080 Qwen2.5 tokens through ten tolerance levels \\
\bottomrule
\end{tabular}
\end{table}

\subsection{Preprocessing and Lead Reduction}
Each lead is centered by subtracting its median, then filtered with a third-order Butterworth high-pass filter (cut-off 0.5 Hz), applied forwards and backwards, which removes baseline wander without phase distortion. The filtered signal serves as the reference for all fidelity measures, so the loss due to the encoding is isolated from the filter's effect.

Because the limb leads are linear combinations of leads I and II, only eight leads are written, and the decoder restores the other four as
\begin{equation}\begin{aligned}\mathrm{III}&=\mathrm{II}-\mathrm{I}, & \mathrm{aVR}&=-\frac{\mathrm{I}+\mathrm{II}}{2},\\ \mathrm{aVL}&=\mathrm{I}-\frac{\mathrm{II}}{2}, & \mathrm{aVF}&=\mathrm{II}-\frac{\mathrm{I}}{2}\end{aligned}\label{eq:limb}\end{equation}

Omitting four channels shortens the text by one-third, but the rebuilt leads inherit the errors of decoded leads I and II, which combine in III, aVL, and aVF. In the stored PTB-XL data, the identities hold to within about 1.5~$\mu$V for almost every record; the 63 records that deviate by more than 2~$\mu$V are excluded in a sensitivity analysis (Section~\ref{sec:main}).

\subsection{Encoding and Decoding}
\subsubsection{Cardiac-cycle-dependent key points}
QRS complexes are located on a multi-lead energy signal $e[n]$, obtained by band-pass filtering all 12 leads (5--15 Hz), taking the Euclidean norm across leads at each sample, and smoothing the result with a 5-sample moving average. Beats are those peaks that have an amplitude of at least 30\% of the 99th percentile of $e[n]$ and occur with at least 0.33 s separation. Then each sample receives a tolerance $\varepsilon[n]$, which depends on the position of the sample relative to the nearest beat; at the default setting, $\varepsilon[n]$ is 0.02 mV for samples between $-$100 and +120 ms from the maximum (QRS complex), 0.03 mV for samples between $-$300 and $-$100 ms (P wave and PR segment), 0.04 mV for samples between +130 and +500 ms (ST segment and T wave), and 0.06 mV for all other samples. The QRS complex has the tightest tolerance because it has a short duration and steep deflections, which would otherwise be the first to be simplified. The windows are fixed offsets around detected peaks rather than around annotated wave boundaries.

Key points are then selected with an RDP variant that uses this per-sample tolerance. The first and last samples are always kept. For two retained samples $a$ and $b$ joined by the straight line $\ell_{ab}[m]$, the algorithm retains the intermediate sample with the largest positive excess
\begin{equation}\delta[m]=\bigl|x[m]-\ell_{ab}[m]\bigr|-\varepsilon[m],\qquad a<m<b\label{eq:excess}\end{equation}

and processes the two resulting sub-segments in the same way until no positive excess remains. The tolerance bounds the piecewise-linear approximation before quantization; it does not bound the final decoded error, which also includes contributions from quantization and lead reconstruction.

\subsubsection{Amplitude alphabet}
\begin{figure}[t]
\centering
\includegraphics[width=\columnwidth]{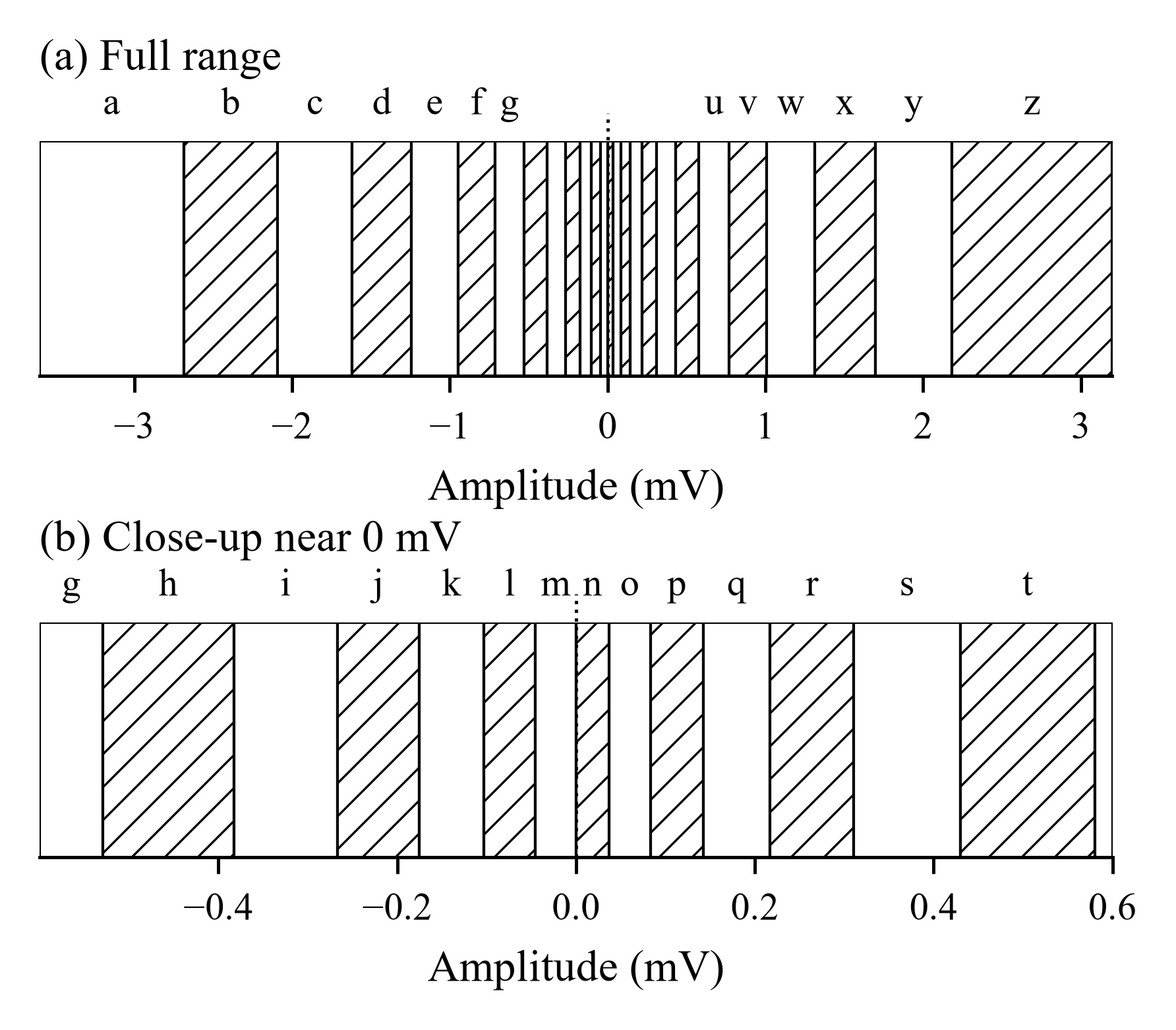}
\caption{The 26-letter amplitude partition, with alternating hatching. (a) Full range; the outer bins a and z extend beyond the plotted range. (b) Close-up of the bins near zero, where they are narrowest. The dotted line marks 0 mV.}
\label{fig:2}
\end{figure}

Key-point voltages are represented by 26 lowercase letters, whose bins are fitted once on training folds 1--8 and then frozen. Letters a--m cover non-positive values and n--z positive ones. The 25 internal boundaries consist of zero and twelve boundaries on each side, spaced by $\mu$-law companding:
\begin{equation}b_j^{\pm}=\pm\frac{R_{\pm}}{\mu}\left[(1+\mu)^{j/13}-1\right],\qquad j=1,\dots,12\label{eq:mulaw}\end{equation}

where $\mu = 20$, and $R_- = 3.441$~mV and $R_+ = 2.795$~mV are the magnitudes of the 0.05th and 99.95th percentiles of the pooled training key-point voltages. The bins are narrowest around the isoelectric line, where the two bins adjacent to zero are about 0.045 and 0.037 mV wide, and widest at the extremes, where the outer bins are open-ended (Fig.~\ref{fig:2}). Each letter maps to the mean training voltage of its bin, minimizing the mean squared quantization error on the training data for the given partition.

Very large deflections would otherwise be clipped by the outer letters. If any key point of a lead falls more than 0.25 mV outside the decoded range, the lead is written at half or quarter scale with the prefix \texttt{x2} or \texttt{x4}, and the decoder multiplies its values by 2 or 4. In the evaluated corpus, 1,653 records have at least one scaled lead.

\subsubsection{Time gaps, serialization and length control}
The gap d between consecutive key points is expressed in samples. The letters A--Z represent gaps of 1--26 samples; a longer gap is written as a run of Z letters followed by the letter for the remainder, and the decoder adds consecutive time letters. Two amplitude letters in a row mean a one-sample step, which is common in steep deflections. Unlike equal-frequency time bins, exact timing ensures that each retained key point is returned to its original sample position.

Symbols are separated by spaces, and each of the eight lines starts with its corresponding lead label (I, II, V1, \dots{}, V6). The unmodified Qwen2.5 tokenizer \cite{r21} is used to measure length. With a 4,096-token context, the signal text can contain at most 4,080 tokens; the remaining 16 tokens cannot hold instructions or answers, so a downstream study must count the complete input of the fully assembled context. A record that is too long at a given level is re-encoded at the next of ten tolerance levels (level 9 uses 0.15, 0.20, 0.30, and 0.50 mV for the four segment types); if it is still too long at level 9, building the dataset stops with an explicit error.

\subsubsection{Decoding and a worked example}
Decoding does the opposite: amplitude letters are transformed to codebook values, and any gain factor is applied; time letters are added together to form sample positions; values between key points are linearly interpolated; and Eq.~\eqref{eq:limb} restores the limb leads. No neural decoder is involved. Re-encoding the decoded key points produces the original string, which verifies the codec but not the original voltages.

\begin{figure}[t]
\centering
\includegraphics[width=\columnwidth]{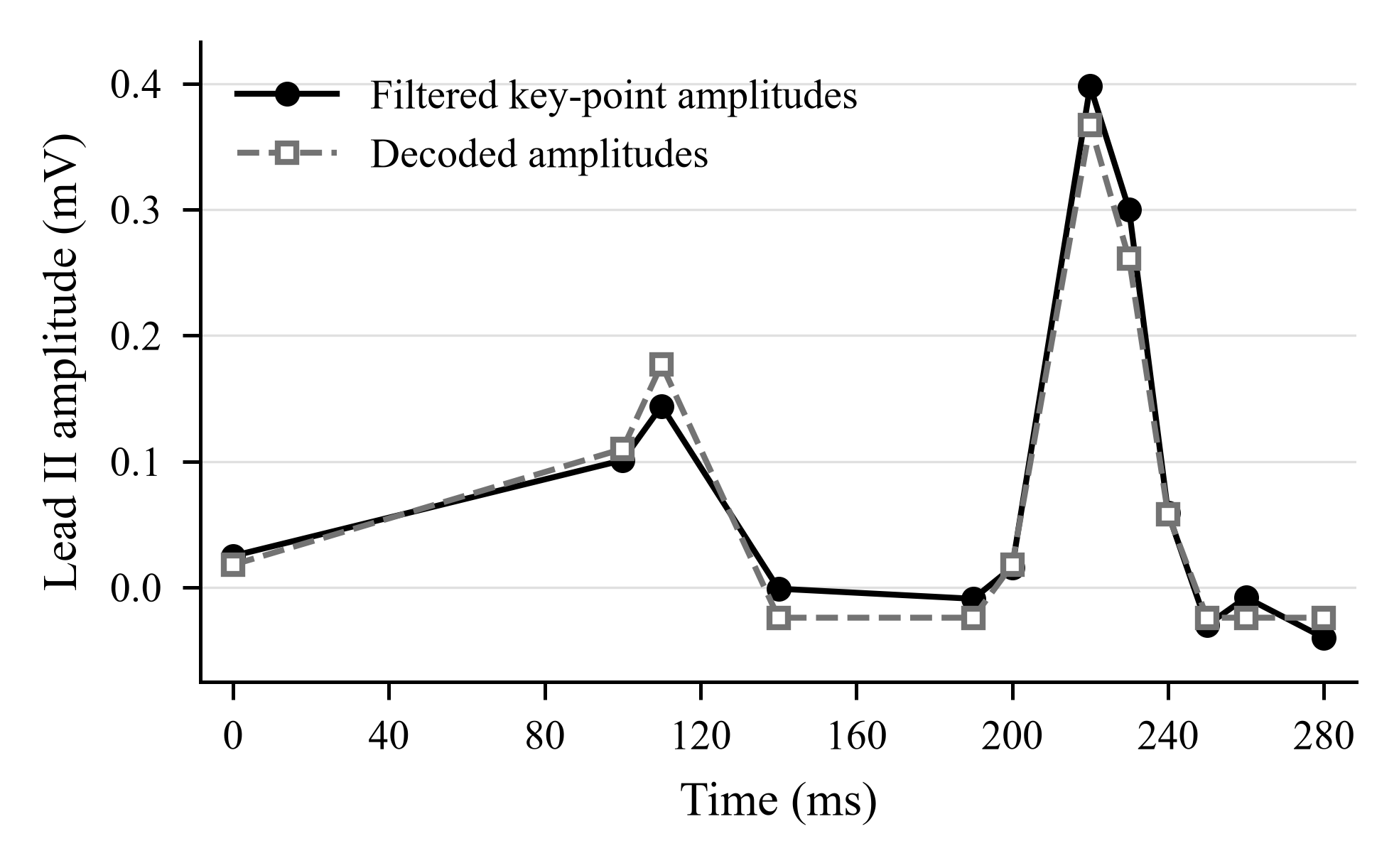}
\caption{Filtered input (solid) and decoded (dashed) amplitudes of the key points in Table 3, covering 0--280 ms. Connecting lines only guide the eye; the example illustrates quantization and is not a diagnostic validation.}
\label{fig:3}
\end{figure}

Table~\ref{tab:3} traces the first 12 key points of lead II in PTB-XL record 1, a training-fold record; the 1,000 samples of this lead are represented by 230 key points and 394 symbols. The first key point (0.025 mV) is encoded as \texttt{n}, which decodes to 0.018 mV. It is followed by a 10-sample interval, written as \texttt{J}. The R-wave peak at sample 22 (220 ms; 0.398 mV) is written as \texttt{s} and decoded as 0.367 mV; key points 7--11 are one sample apart, so no time letters appear between them. Reading Table~\ref{tab:3} row by row yields the opening of the lead II line \texttt{n J p q C m E m n B s r o m m B m}, and the input and decoded amplitudes are compared in Fig.~\ref{fig:3}.

\begin{table}[t]
\caption{Worked example: the first twelve key points of lead II in PTB-XL record 1. Every key point contributes its amplitude letter and then the time letter of the following gap, which is omitted when the next key point is one sample away.}
\label{tab:3}
\centering\footnotesize
\setlength{\tabcolsep}{3.5pt}
\begin{tabular}{@{}rrrcrrc@{}}
\toprule
Point & Sample & Input (mV) & Letter & Decoded (mV) & Gap & Time letter\\
\midrule
1 & 0 & 0.025 & n & 0.018 & 10 & J \\
2 & 10 & 0.101 & p & 0.110 & 1 & -- \\
3 & 11 & 0.144 & q & 0.177 & 3 & C \\
4 & 14 & $-$0.001 & m & $-$0.024 & 5 & E \\
5 & 19 & $-$0.009 & m & $-$0.024 & 1 & -- \\
6 & 20 & 0.016 & n & 0.018 & 2 & B \\
7 & 22 & 0.398 & s & 0.367 & 1 & -- \\
8 & 23 & 0.300 & r & 0.261 & 1 & -- \\
9 & 24 & 0.059 & o & 0.058 & 1 & -- \\
10 & 25 & $-$0.030 & m & $-$0.024 & 1 & -- \\
11 & 26 & $-$0.008 & m & $-$0.024 & 2 & B \\
12 & 28 & $-$0.040 & m & $-$0.024 & 1 & -- \\
\bottomrule
\end{tabular}
\end{table}

\subsection{Computational Cost}
Encoding is dominated by key-point selection and token counting; the recursive RDP search is, at worst, quadratic in the number of samples in a lead, and only records that exceed the budget are encoded more than once. Decoding needs only table look-ups and linear interpolation. Both arms of the downstream experiment use the same network and take about 219 s per run on a single NVIDIA T4 GPU. The benefit of shorter text lies outside the classifier: because attention cost grows with input length, fewer tokens make language-model training and inference cheaper.

\section{Experimental Setup}\label{sec:setup}
\subsection{Dataset}
The experiments use PTB-XL version 1.0.3 \cite{r22}, a public database of 21,799 ten-second, twelve-lead ECGs from 18,869 patients. In the 100-Hz release, every record is a 12 $\times$ 1,000 matrix in millivolts. Records that carried at least one of the 44 diagnostic statements of the SCP-ECG standard were retained, resulting in 21,388 records from 18,617 patients. They were labeled with the five diagnostic superclasses of PTB-XL, namely normal ECG (NORM), myocardial infarction (MI), ST/T change (STTC), conduction disturbance (CD), and hypertrophy (HYP), using the standard aggregation at any likelihood; a record may therefore carry several labels.

The recommended patient-wise split assigns folds 1--8 to training (17,084 records), fold 9 to validation (2,146 records), and fold 10 to testing (2,158 records from 1,877 patients), with no patient in more than one split. The test fold contains 963 NORM, 550 MI, 521 STTC, 496 CD, and 262 HYP positives. All arms share the same records, labels, and preprocessing, so differences between them cannot be attributed to the data.

\subsection{Evaluation Metrics}
\textit{Codec integrity.} Every record was decoded and re-encoded; the regenerated lead strings were compared character by character with the stored ones, and token counts were recomputed.

\textit{Signal fidelity.} The decoded signal $\hat{x}$ was compared with the filtered reference $\tilde{x}$ through the Pearson correlation coefficient r, averaged over the twelve leads of each record, and through the normalized percentage root-mean-square difference
\begin{equation}\mathrm{PRDN}=100\times\sqrt{\frac{\sum_{\ell}\sum_{n}\left(\hat{x}_\ell[n]-\tilde{x}_\ell[n]\right)^2}{\sum_{\ell}\sum_{n}\left(\tilde{x}_\ell[n]-\bar{\tilde{x}}_\ell\right)^2}}\label{eq:prdn}\end{equation}

where the barred term is the mean of reference lead $\ell$. PRDN was computed over all 12 leads, the 8 stored leads, and the 4 rebuilt leads. Normalizing by the variance, rather than by the reference energy, makes it independent of any constant offset in the reference.

\textit{Diagnostic information.} Classifier quality was summarized by the area under the receiver operating characteristic curve (AUROC) for each class and by its macro-average, which weights the five classes equally and requires no decision threshold. Micro-AUROC, macro-averaged average precision (AP), and macro-F1 are secondary metrics. The primary estimand is the paired difference
\begin{equation}\Delta=\frac{1}{S}\sum_{s=1}^{S}\left(\mathrm{AUROC}_{\mathrm{macro}}^{\mathrm{dec},s}-\mathrm{AUROC}_{\mathrm{macro}}^{\mathrm{ref},s}\right)\label{eq:delta}\end{equation}

the mean over S = 5 seeds of the test macro-AUROC of the decoded arm minus that of the reference arm.

\subsection{Baselines and Paired Design}
The main comparison is internal: the reference arm trains a classifier on the filtered signal, and the decoded arm trains the same classifier on the signal decoded from ECG-26c text. To isolate each design component, all records were also encoded using the same codebook in six fixed variants: every sample quantized without key-point selection (quantization only), a uniform RDP tolerance of 0.02 mV, and tolerance levels of 0, 3, 6, and 9 applied without the budget rule. No published ECG text method was re-implemented; those systems are compared by design only (Table~\ref{tab:2}).

The classifier is a one-dimensional residual convolutional neural network (CNN): a stem convolution (kernel 7, stride 2, 16 channels), four residual blocks widening to 32, 64, 64 and 128 channels (kernel 5, batch normalization, ReLU), concatenated global average and max pooling, dropout of 0.2 and a five-unit output layer, with 221,477 parameters in total. It was chosen from 12 configurations using only the reference arm, so any advantage from model selection lies with the reference. Training minimized binary cross-entropy with AdamW (weight decay 0.01) under a one-cycle learning-rate schedule peaking at $3 \times 10^{-3}$, using batches of 64 for 30 epochs without augmentation. The checkpoint with the highest validation macro-AUROC was retained, and per-class thresholds that maximized validation F1 were applied unchanged to the test fold. For each seed (0--4), the two arms shared their initial weights and the mini-batch order.

Uncertainty was quantified using 10,000 patient-cluster bootstrap replicates of the test fold, in which patients, rather than records, are resampled and both arms share the same draws \cite{r23,r24}. The decoded arm is declared non-inferior if the lower bound of the two-sided 95\% interval for $\Delta$ lies above $-$0.010, following the paired-AUC framework of Liu et al. \cite{r25}. This margin is an analytic tolerance, not a clinical acceptance threshold. It was recorded on 30 September 2026, and the analysis plan was frozen on 14 October 2026 before any model for this experiment was trained; however, earlier encoder development had used fold 10 (Section~\ref{sec:limits}). Secondary metrics are not corrected for multiple comparisons, and the fixed-variant results are exploratory.

\subsection{Experimental Environment}
The experiments used Python 3.13, NumPy 2.4, SciPy 1.17, pandas 3.0, scikit-learn 1.9, and PyTorch 2.11 with CUDA 13.0 on an NVIDIA Tesla T4 GPU.

\section{Results and Discussion}\label{sec:results}
\subsection{Main Results}\label{sec:main}
\textit{Codec integrity and text length.} All 171,104 stored lead strings (21,388 records $\times$ 8 leads) were reproduced exactly, and all ten fold files were rebuilt from the raw PTB-XL files with matching contents. The average (median) number of tokens per record was 2,866 (2,795), with a 95th percentile of 3,880 and a maximum of 4,080. No record was truncated, and 94.1\% of records were written at level 0. The symbol count is close to the token count, with 1.01 symbols per space-separated token.

\begin{figure}[t]
\centering
\includegraphics[width=\columnwidth]{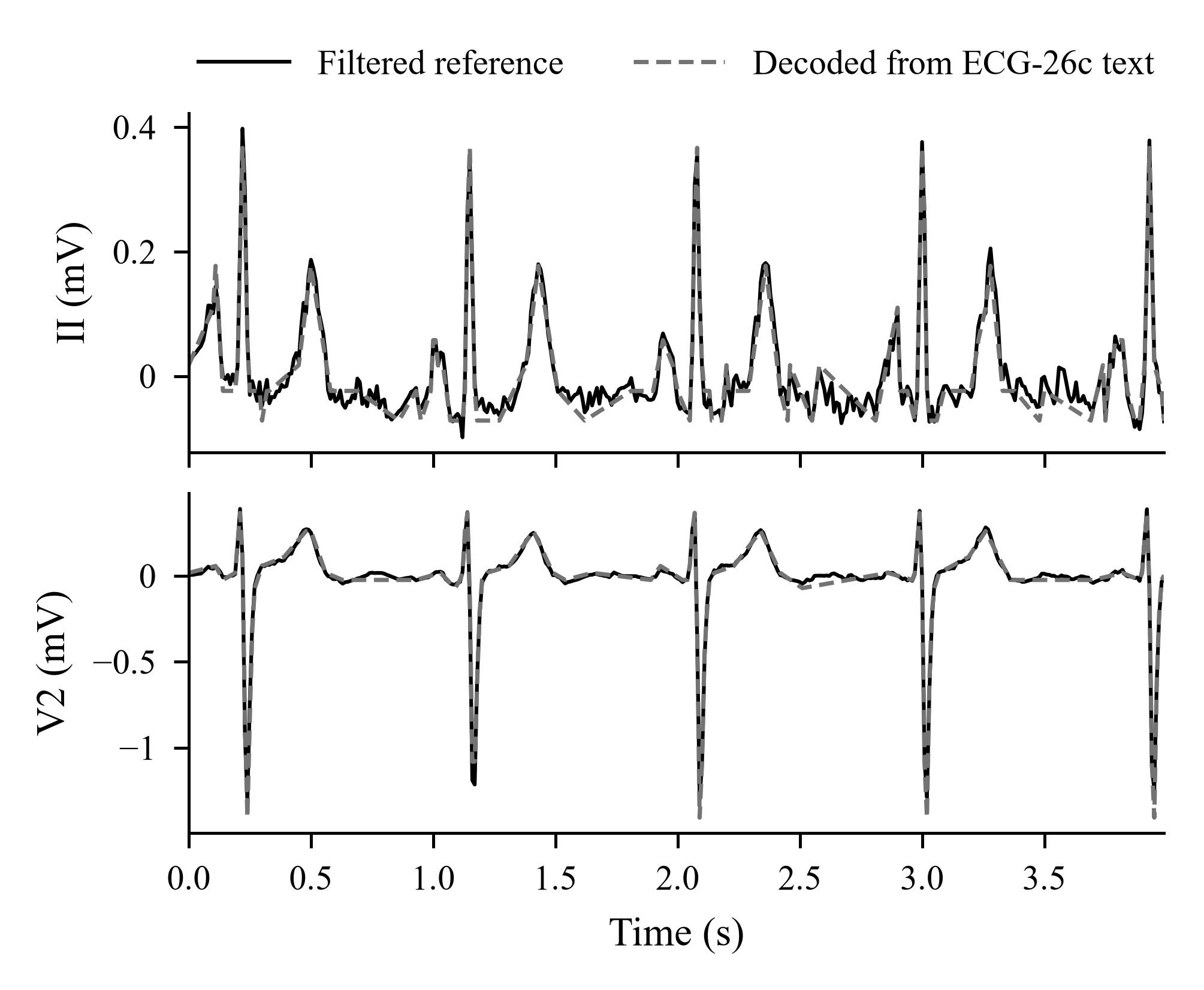}
\caption{The first four seconds of leads II and V2 in PTB-XL record 1: filtered reference (solid line) and signal decoded from ECG-26c text (dashed line).}
\label{fig:4}
\end{figure}

\textit{Signal fidelity.} The average r value between the decoded signals and the filtered reference was 0.982. The median PRDN was 12.7\% for all 12 leads, 11.4\% for the stored leads, and 20.6\% for the rebuilt leads, indicating that errors in leads I and II are compounded in the derived leads. A typical reconstruction is depicted in Fig.~\ref{fig:4}; the amplitudes and shapes of the QRS complexes and T waves are retained, and the small baseline fluctuations are smoothed.

\textit{Diagnostic information.} The two arms in the test fold are compared in Table~\ref{tab:4}. The macro-AUROC was 0.9167 $\pm$ 0.0014 (mean $\pm$ SD over seeds) for the reference and 0.9157 $\pm$ 0.0010 for the decoded signal. The lower limit of the 95\% interval for the paired difference is well above the pre-specified margin of $-$0.010, so the decoded arm is non-inferior, and the interval includes zero. The seed-to-seed standard deviation of the paired difference (0.0014) exceeded the mean difference itself. The CNN is compact rather than state-of-the-art: deeper PTB-XL benchmark models achieve superclass macro-AUROC values of roughly 0.92--0.93 \cite{r19}.

\begin{table*}[t]
\caption{Test-fold metrics of the CNN (mean $\pm$ SD over five seeds; thresholds selected on the validation fold) and paired differences with 95\% patient-cluster bootstrap intervals. The higher mean in each row is shown in bold.}
\label{tab:4}
\centering\footnotesize
\setlength{\tabcolsep}{8pt}
\begin{tabular}{@{}lccc@{}}
\toprule
Metric & Reference & Decoded & Paired $\Delta$ [95\% interval]\\
\midrule
Macro-AUROC (primary) & \textbf{0.9167 $\pm$ 0.0014} & 0.9157 $\pm$ 0.0010 & $-$0.0010 [$-$0.0028, 0.0007] \\
Micro-AUROC & \textbf{0.9281 $\pm$ 0.0027} & 0.9253 $\pm$ 0.0015 & $-$0.0029 [$-$0.0044, $-$0.0013] \\
Macro-AP & 0.8025 $\pm$ 0.0044 & \textbf{0.8026 $\pm$ 0.0033} & 0.0001 [$-$0.0036, 0.0039] \\
Macro-F1 & \textbf{0.7304 $\pm$ 0.0027} & 0.7276 $\pm$ 0.0029 & $-$0.0027 [$-$0.0081, 0.0026] \\
AUROC, NORM & 0.9393 $\pm$ 0.0041 & \textbf{0.9408 $\pm$ 0.0028} & 0.0015 [$-$0.0006, 0.0036] \\
AUROC, MI & \textbf{0.9115 $\pm$ 0.0074} & 0.9074 $\pm$ 0.0024 & $-$0.0041 [$-$0.0081, $-$0.0002] \\
AUROC, STTC & \textbf{0.9299 $\pm$ 0.0031} & 0.9291 $\pm$ 0.0008 & $-$0.0008 [$-$0.0038, 0.0022] \\
AUROC, CD & \textbf{0.9101 $\pm$ 0.0057} & 0.9091 $\pm$ 0.0063 & $-$0.0011 [$-$0.0051, 0.0028] \\
AUROC, HYP & \textbf{0.8929 $\pm$ 0.0030} & 0.8922 $\pm$ 0.0049 & $-$0.0007 [$-$0.0050, 0.0036] \\
\bottomrule
\end{tabular}
\end{table*}

Non-inferiority on the primary metric does not mean identical behavior. Micro-AUROC and MI AUROC decreased with intervals that exclude zero, and the mean per-seed MI F1 score fell by 0.013. Averaging the five seeds into an ensemble also revealed a small macro-AUROC loss ($\Delta$ = $-$0.0023; 95\% interval $-$0.0042 to $-$0.0003). Even so, the thresholded ensemble predictions of the two arms agreed closely, with Cohen's $\kappa$ from 0.857 (HYP) to 0.934 (NORM), and 83.4\% of test records received the same label set. The conclusion held in two pre-specified subsets: $\Delta$ = $-$0.0007 ($-$0.0026 to 0.0011) for the 2,011 test records encoded at level 0, and $\Delta$ = $-$0.0010 ($-$0.0029 to 0.0008) for the 2,151 records whose raw limb leads satisfy Eq.~\eqref{eq:limb} within 2~$\mu$V.

\subsection{Ablation Study}
\begin{figure}[t]
\centering
\includegraphics[width=\columnwidth]{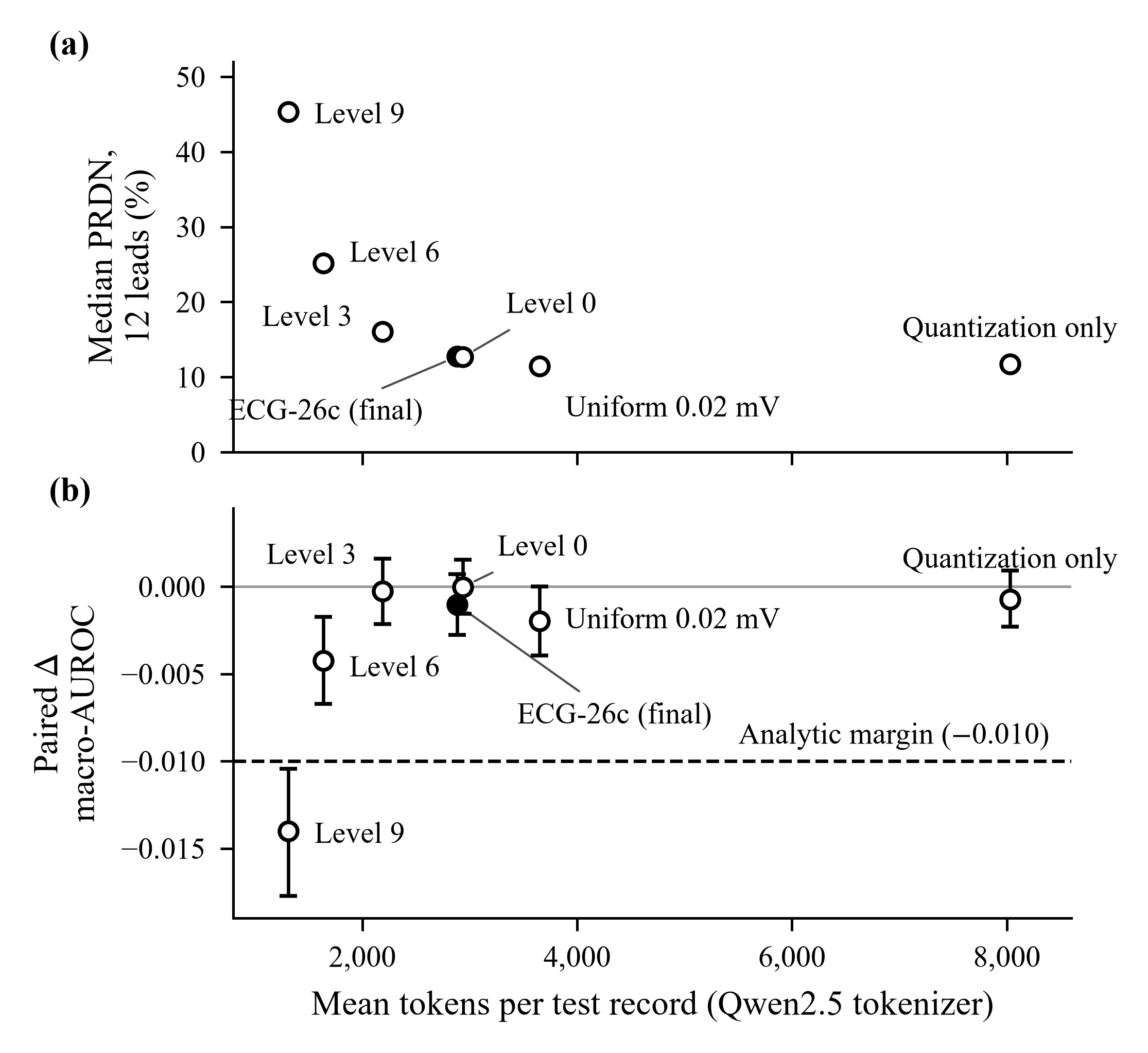}
\caption{Compression trade-off on the test fold. (a) Median PRDN over the twelve leads. (b) Mean paired change in macro-AUROC with 95\% patient-cluster bootstrap intervals. The filled marker denotes the final ECG-26c encoding, and the dashed line marks the analytic margin of $-$0.010.}
\label{fig:5}
\end{figure}

In Table~\ref{tab:5} and Fig.~\ref{fig:5}, the contributions of amplitude quantization, key-point selection, and the budget rule are separated. Quantization without key-point selection, which discards no samples, gives a median PRDN of 11.7\% at a cost of 8,029 tokens per record. A uniform RDP tolerance of 0.02 mV reduces the number of tokens by more than half with similar fidelity, but the longest records still need more than 7,000 tokens. Cycle-dependent tolerances at level 0 save a further 19\% with a modest increase in error, but about 6\% of records still exceed 4,080 tokens. The budget rule removes this tail: these records are re-encoded at coarser levels, which caps the maximum at 4,080 tokens per record without changing the median PRDN.

\begin{table*}[t]
\caption{Compression, reconstruction, and paired diagnostic change for each encoding variant. Token and PRDN values cover all 21,388 records; $\Delta$ macro-AUROC refers to the test fold, with 95\% patient-cluster bootstrap intervals. Savings are relative to quantization only.}
\label{tab:5}
\centering\footnotesize
\begin{tabular}{@{}lrrrcc@{}}
\toprule
Variant & Mean tokens & Max tokens & Saving & Median PRDN (\%) 12 / 8 / 4 leads & $\Delta$ macro-AUROC [95\% interval]\\
\midrule
Quantization only & 8,029 & 8,043 & 0\% & 11.7 / 10.6 / 18.7 & $-$0.0007 [$-$0.0023, 0.0009] \\
Uniform 0.02 mV & 3,611 & 7,728 & 55.0\% & 11.4 / 10.4 / 18.2 & $-$0.0020 [$-$0.0040, 0.0000] \\
Level 0 (fixed) & 2,912 & 7,462 & 63.7\% & 12.7 / 11.4 / 20.5 & 0.0000 [$-$0.0016, 0.0016] \\
\textbf{ECG-26c (final)} & 2,866 & 4,080 & 64.3\% & 12.7 / 11.4 / 20.6 & $-$0.0010 [$-$0.0028, 0.0007] \\
Level 3 (fixed) & 2,177 & 6,702 & 72.9\% & 16.1 / 14.4 / 26.0 & $-$0.0002 [$-$0.0021, 0.0016] \\
Level 6 (fixed) & 1,637 & 5,372 & 79.6\% & 25.3 / 22.3 / 42.1 & $-$0.0042 [$-$0.0067, $-$0.0017] \\
Level 9 (fixed) & 1,313 & 3,416 & 83.7\% & 44.8 / 39.4 / 73.1 & $-$0.0140 [$-$0.0177, $-$0.0104] \\
\bottomrule
\end{tabular}
\end{table*}

Coarser fixed levels trace the trade-off. At level 3 the error rises noticeably, yet the change in macro-AUROC cannot be distinguished from zero. A detectable loss appears at level 6, while level 9 crosses the margin. Level 6 thus separates a statistical loss from a violation of non-inferiority: its whole interval lies below zero, so information is lost, but not enough to breach the chosen margin. Because these comparisons use the test fold, they describe the trade-off and should not be used to select a new operating level.

\subsection{Discussion}
The small losses were all in myocardial infarction. A possible explanation, not directly tested here, is that brief deflections, such as pathological ``Q'' waves (only a few samples at 100 Hz), are small and can fall within the tolerance or a single amplitude bin. A wave-level audit is consistent with that, showing that among detectable waves, 2.0\% of Q waves were missing after decoding, while 0.08\% of R waves and 0.04\% of T waves were missing.

\textit{Reconstruction error as a proxy.} Record-level error did not predict how much a record's predictions changed: the Spearman correlation between record-level PRDN and the mean absolute change in predicted probability was $-$0.08 (n = 2,158), statistically detectable but negligible. Together with the absence of loss at level 3, this agrees with the long-standing argument that PRD-type measures do not capture diagnostic distortion \cite{r18}. Text representations of ECGs should therefore be judged by direct tests of diagnostic content rather than by fidelity thresholds.

\textit{Compression metadata as a shortcut.} Coarser levels were needed more often for abnormal records: 7.4--10.8\% of records in the four disease classes required a level above 0, compared with 2.45\% of NORM records. Fast rhythms and noisy recordings generate more key points, so the pattern is expected, but the level and the gain markers are visible in the text, and a language model could exploit them as shortcuts. Downstream experiments should withhold the level from the prompt or stratify their results by level.

\textit{Implications for language-model studies.} The token counts reported here depend on the Qwen2.5 tokenizer and cannot be compared directly with those of ECG-Byte, whose vocabulary is trained on ECG strings \cite{r4}, or of ECG-aBcDe, which does not report token counts \cite{r5}. The distinctive element of this study is its evaluation, which measures information loss before any language model is involved. A non-inferior CNN, which receives an interpolated waveform, shows that the information is present in the text, not that a language model can extract it from letters. In an earlier project audit, an order-free classifier using only per-lead symbol frequencies reached a test macro-AUROC of about 0.87; future language-model results should beat such baselines to show that they exploit the temporal structure of the text.

\subsection{Limitations}\label{sec:limits}
The main limitation is that the test fold was partly exposed during development. The tolerances, filter type, and quantizer type were selected on the first 100 labeled records of fold 10. The previous CNN validation also included fold 10, and the label-based cross-validation used the first 100 labeled records of every fold. The frozen paired protocol controls the comparison between the arms but cannot undo these earlier decisions, so the result is a structured re-estimation that must be confirmed on independent data.

The evidence is for one dataset, five broad superclasses, and a compact CNN family; more powerful classifiers could show losses that this model does not, and the impact of the 0.5-Hz filter itself is not quantified. The rebuilt limb leads have nearly twice the relative error of the original leads, and the smallest beat distance is 0.33 s (180 bpm), which could merge beats in faster rhythms. Lastly, the outer amplitude bins are open-ended, so values in them can be coded with errors of up to 0.7 mV, and 16 records exceed the coded range even at quarter scale.

\section{Conclusion}\label{sec:conclusion}
This paper showed how a twelve-lead ECG can be written as compact, decodable text and how the information retained in that text can be measured without a language model. ECG-26c writes eight leads as amplitude letters and exact time gaps, selects key points with heartbeat-dependent tolerances, and adapts its tolerance level so that every PTB-XL record fits within 4,080 Qwen2.5 tokens. Across 21,388 records, the codec was exactly consistent, and the mean token count was roughly one-third of that for a per-sample transcription. In a paired five-seed CNN experiment, the decoded signals were non-inferior on macro-AUROC; the losses in micro-AUROC and in the MI class were small but detectable and stayed within the margin. At the same time, the ablation indicated that reconstruction error is a poor predictor of diagnostic loss. The main lesson for practitioners is that representation fidelity, diagnostic sufficiency, and language-model capability are distinct properties and therefore require different tests. Three directions follow. First, the design choices should be repeated on the training data only and checked on an independent dataset. Second, the representation should be improved where it is weakest (the open outer amplitude bins, small deflections such as Q waves, and fast rhythms). Third, language models should be fine-tuned on ECG-26c text with the full context counted, compared against order-free symbol-frequency baselines, and tested for shortcuts through compression levels and gain markers. These are only computational comparisons and would not be sufficient for clinical use.

\section*{Acknowledgment}
During the preparation of this paper, the authors used Anthropic Claude (Opus 5.5) for language editing, restructuring the text according to the book and conference templates, checking and formatting references, and preparing figures from existing project data. No experimental results were generated with these tools; all reported values come from the recorded project experiments. After using these tools, the authors reviewed and edited the content as needed and took full responsibility for the publication's content.

\section*{Data Availability}
PTB-XL version 1.0.3 is publicly available through PhysioNet at \url{https://physionet.org/content/ptb-xl/1.0.3/} under a Creative Commons Attribution 4.0 license \cite{r22}. The ECG-26c encoder, standalone decoder, codebook, derived fold files, reproduction notebook and evaluation reports are available at \url{https://github.com/ekabullar/-ecg26c-llm}, subject to the repository's access and licensing terms. The study uses an existing de-identified dataset and involves no newly collected participant data.

\end{document}